# Evaluation of Machine Translation Based on Semantic Dependencies and Keywords

Kewei Yuan[1], Qiurong Zhao[2], Yang Xu[1], Xiao Zhang[1], Huansheng Ning[1]

1) school of Computer & Communication Engineering, University of Science and Technology Beijing, Beijing, China

2) School of foreign languages, Beijing University of science and technology, Beijing, China

**Abstract**——In view of the fact that most of the existing machine translation evaluation algorithms only consider the lexical and syntactic information, but ignore the deep semantic information contained in the sentence, this paper proposes a computational method for evaluating the semantic correctness of machine translations based on reference translations and incorporating semantic dependencies and sentence keyword information. Use the language technology platform developed by the Social Computing and Information Retrieval Research Center of Harbin Institute of Technology to conduct semantic dependency analysis and keyword analysis on sentences, and obtain semantic dependency graphs, keywords, and weight information corresponding to keywords. It includes all word information with semantic dependencies in the sentence and keyword information that affects semantic information. Construct semantic association pairs including word and dependency multi-features. The key semantics of the sentence cannot be highlighted in the semantic information extracted through semantic dependence, resulting in vague semantics analysis. Therefore, the sentence keyword information is also included in the scope of machine translation semantic evaluation. To achieve a comprehensive and in-depth evaluation of the semantic correctness of sentences, the experimental results show that the accuracy of the evaluation algorithm has been improved compared with similar methods, and it can more accurately measure the semantic correctness of machine translation.

# 1 Introduction

Machine translation (MT) first appeared in the 1950s. After more than 70 years of development, machine translation methods have been continuously expanded and deepened, and the quality of machine translation has been improved by leaps and bounds. From early rule-based machine translation to neural network machine translation characterized by deep learning models and implemented in the 21st century. The translation mechanisms of the existing machine translation APIs on the market are different, and the output results of English-to-Chinese translation are also different. Evaluation of these machine translation results is also a necessary link in the development of machine translation. How to comprehensively evaluate the quality of machine translation is a problem to be solved by machine translation.

Previous studies on machine translation evaluation (MTE) are divided into manual evaluation methods and automatic evaluation methods. Manual evaluation can directly reflect the user's satisfaction with the quality of machine translation, (research flaw), but the evaluation process takes a lot of manpower and financial resources, and there are personal subjective factors, so the consistency of evaluation results cannot be guaranteed. The automatic evaluation method that came into being is mainly based on the sentence similarity between the machine translation and the reference translation, and most of them focus on the lexical and syntactic levels. For example, in the algorithms BLEU, TER, METEOR, etc. from the perspective of vocabulary, these methods of evaluating lexical similarity cannot fully reflect the semantic similarity of sentences. Secondly, to capture the semantics or the semantics of text fragments, entity naming knowledge, and semantic role labeling are used in machine translation evaluation research, but these features cannot reflect the second-level semantic relationship of sentences.

This project research can evaluate the semantic dependence of existing translation APIs in terms of sentence semantics, can effectively solve the problem of screening the best translation, and can also help machine translation to a certain extent. Therefore, this research needs to solve how to comprehensively evaluate the semantic correctness of machine translation from the perspective of semantics.

Semantic analysis is a dependency-based language structure analysis [i](Hudson and Hudson (2007)), whose purpose is to figure out the deep semantic structure of sentences. Semantic dependency includes the main semantic role, each semantic role corresponding to a nested relationship and anti-relationship; event relationship, describing the relationship between two events; semantic dependency mark, marking the tone of speech and other dependency information. The semantics of a sentence is closely related to the structure of a sentence. Semantic dependency analysis is a more complete and comprehensive description of the semantic information of a sentence, and it is a further deepening of syntax and semantics. This paper proposes to add a calculation method for the keyword similarity of two sentences based on the semantic dependence and word semantic similarity calculation method, taking into account the semantic features of multiple levels such as the deep semantic relationship in the sentence, word semantics, and sentence keywords.

# 2 Related Works

## 2.1 Research progress in the automatic evaluation of machine translation

In terms of lexical similarity, Su, Wu, and Chang (1992)[ii] introduced word error rate (WER) into machine translation evaluation, by Tillmann et al. (1997)[iii] proposed position-independent word error rate (PER) to ignore word order when matching; Turian, Shea, and Melamed (2006)[iv] proposed the TER algorithm with block movement as an editing step; Papineni et al. (2002)[v] proposed the first commonly used evaluation method BLEU in the industry. In the BLEU measurement, the weight of n-grams is usually set to a uniform weight. Banerjee and Lavie (2005)[vi] proposed a new evaluation metric, METEOR, based on flexible single-word matching, single-word precision, and single-word recall, including simple morphological variables for matching by identical words and synonyms.

Although some lexical similarity-based methods mentioned above take into account linguistic information, such as synonyms, lexical similarity methods mainly focus on the precise matching of superficial words in machine translation and seldom consider syntactic information, and a good translation is and It is semantically unreasonable for reference translations to agree only on lexical choices. Reflecting semantic similarity is not only superficial lexical similarity but also needs similarity based on semantic structure. Li, Gong, and Zhou (2012)[vii] use noun phrase and verb phrase similarity, and Echizen-ya and Araki (2010)[viii] perform an automatic evaluation of machine translation using only noun phrase groups. To capture the semantic equivalence of text fragments, entity naming is used in machine translation evaluation, and Koehn et al. (2007)[ix] exploit the open NLP toolbox to detect entity naming. From the perspective of entity naming, the inefficiency leads to a decrease in the fluency and comprehensiveness of the evaluation. But from the perspective of improving the similarity with human evaluation, the entity naming recognition feature has almost the largest contribution compared with other features. Some researchers use semantic roles as linguistic features in machine translation evaluation, and Giménez and Màrquez (2008)[x] cite semantic roles to account for arguments and adjuncts that appear in candidate and reference translations. Lo, Tumuluru and Wu (2012)[xi] weight different types of semantic roles according to their relative importance. In general, semantic roles play a role in the syntactic structure of a sentence. Some researchers use deep learning and neural network models for machine translation evaluation. For example, Guzmán et al. (2017)[xii] use MTE's neural network for pairwise modeling to integrate syntactic and semantic information into Among NNs, the best translation is selected. But this approach is too complex.

## 2.2 Research Progress on Semantic Dependency Analysis of Chinese Sentences

Liu (2009)[xiii] proposed that in "dependency grammar", each group of two words constitutes a relationship, one word is the dominant word, and the other word is the subordinate word. The so-called "dependency grammar" reveals its syntactic structure by analyzing the dependency relationship between components in a language unit. Wimmer (1980)[xiv] developed the concept of dependency distance (DD), which represents the linear distance between a dominant word and a subordinate word. Liu, Xu, and Liang (2017)[xv] proposed a DD calculation method: the linear difference between the position of the dominant word and the subordinate word in the sentence. Jiang, Fan, and Wang (2021)[xvi] used dependency distance and dependency direction to verify that there is a significant difference between the translated text and the native language text of the target language.

Chinese semantic analysis, including partial semantic dependency annotation, semantic role standard, shallow semantic analysis, etc. Yuan (2008)[xvii] defined 23 arguments and marked the news corpus. Xue and Palmer (2005)[xviii] used the Chinese Proposition Bank to annotate the semantic roles of Chinese verbs. At present, there is no practical algorithm for semantic dependency analysis, and the related algorithms are dependency syntax analysis and related algorithms for semantic role labeling based on syntax analysis. Therefore, it is necessary to go further in the study of Chinese semantic dependency analysis.

## 2.3 Semantic Dependency Analysis

This article uses the semantic dependency analysis service provided by the Language Technology Platform (LTP) of the Harbin Institute of Technology. The language technology platform is a complete set of open Chinese natural language processing systems developed by the Social Computing and Information Retrieval Center of Harbin Institute of Technology for ten years. The system provides Chinese word segmentation and part-of-speech tagging, Dependency Syntax Analysis, Semantic Dependency Analysis, Keyword Extraction, and other services, all services have been deployed to the Xunfei Open Platform. Semantic dependency analysis represents the meaning of a sentence through a set of dependent word pairs and their corresponding relationships. This process is not affected by sentence changes, such as "他把杯子打碎了。" and "杯子被他打碎了。" These two sentences have the same semantics, but different syntax. Literature 2016 defines a Chinese semantic dependency scheme based on Chinese-specific linguistic knowledge to represent the semantics of sentences in a graphical format. The meaning of a sentence includes the meaning of semantic units and their combinations, including sentence semantic relations and additions.

# 3 Research methods

Due to the existing research on Chinese semantic analysis, most of them is shallow semantic analysis, and there is a lack of deep semantic research. Semantic dependency analysis can clarify the deep semantic structure of a sentence. It is necessary to consider not only the semantics between the words that make up the sentence but also the grammatical structure of the sentence and the impact of keywords on the semantics of the sentence. Provides the semantic dependency analysis interface and keyword analysis interface of the language technology platform developed by Harbin Institute of Technology to obtain the semantic dependency information and keyword information in the sentence, and obtain the sentence-level similarity between machine translation and reference translation by calculating word similarity.

## 3.1 Computation of Semantic Similarity Based on Semantic Dependency

After semantic dependency analysis, this paper proposes to construct a semantic association pair $(P, R)$, where $P$ represents a word pair $(W_1, W_2)$, $W_1$ is a dependent word, $W_2$ is a dependent word, and $R$ is the dependency relationship between word pairs. Che et al. (2016)[xix] mentions that the meaning of a sentence includes the meaning of semantic units and their combinations, including semantic relations and additions. Semantic attachments refer to markings on semantic units. Based on the combined semantic relationship as the starting point, this paper calculates the semantic similarity under each semantic relationship in the machine translation sentence and the reference translation sentence and combines them, and finally obtains the semantic similarity between the machine translation and the reference translation sentence.

1) Assuming that the calculation reference translation sentence is $S_1$, and the machine translation sentence is $S_2$. Call the semantic dependency analysis interface of the semantic technology platform to obtain the semantic dependency analysis information, obtain the dependency relationship graph (which contains several semantic association pairs), and then process the structured data and Remove dependencies on punctuation marks that have no real semantics (such as "mPunc").

2) According to the obtained semantic association pairs, the set of dependency types $Rel_1 = \{R_{11}, R_{12}, \ldots, R_{1n}\}$ of the human-translated sentence $S_1$, and the set of dependency types of the machine-translated sentence $S_2$: $Rel_2 = \{R_{21}, R_{22}, \ldots, R_{2m}\}$. Assume that there are n types of dependency relationships in S1, and m types of dependency relationships in $S_2$. The number of words in each relationship type set $Num_1 = \{N_{11}, N_{12}, ..., N_{1n}\}$; $Num_2 = \{N_{21}, N_{22}, ..., N_{2m}\}$. After semantic dependency analysis, the semantic association pair of $S_1$ is $(P_{1ij}, R_{1i})$, where $P_{1ij}$ represents the jth word pair $(W_{1ij1}, W_{1ij2})$ in the $i-th$ relationship, where $i = (1,2,...,n)$ and $j = (1,2,...,m)$. For example, "港口几周后才恢复运行。" After semantic dependency analysis, the semantic association pair of this sentence is obtained ---"[((恢复,港口),Exp),((周,恢复),Time),((运行,恢复),Cont),((几,周),Meas),((后,周),mDepd),((才,恢复),mDepd)]". The semantic association pair "((恢复,港口),Exp))" indicates that "港口" and "恢复" are parties. The semantic dependency type set $Rel = \{Exp, Time, Cont, Meas, mDepd\}$ obtained in this sentence, and the set of associated pairs of each type of relationship is $Num = \{1,1,1,1,2\}$.

3) Referring to Zhu, Runcong Ma and Sun (2016)[xx], they proposed a word similarity calculation method based on CNKI and Cilin, which uses a fusion calculation method to expand the vocabulary coverage and improve the rationality of the calculation results. The calculated similarity between two words is denoted as $Sw(c_1, c_2)$.

4) The similarity of word pairs is recorded as

$$Sp(P_{1ij}, P_{2kt}) = \frac{w(W_{1ij1}, W_{2kt1}) + Sw(W_{1ij2}, W_{2kt2})}{2}$$

In the formula, $Sw(W_{1ij1}, W_{2kt1})$ represents the word similarity between the dependent word in the $j-th$ word pair in the $i-th$ relationship in $S_1$ and the dependent word in the $t-th$ word pair in the $k-th$ relationship in $S_2$; $Sw(W_{1ij2}, W_{2kt2})$ represents the word similarity between the dependent word in the $j-th$ word pair in the $i-th$ relationship in $S_1$ and the dependent word in the $t-th$ word pair in the $k-th$ relationship in $S_2$.

5) Taking $S_1$ as a reference, if $R_{2t} = R_{1i}$ exists, and $R_{2t}$ represents the $t-th$ dependency relationship in the dependency relationship set of $S_2$ then the similarity of the $j-th$ word pair in the $i-th$ relationship is

$$S_{ij} = \max(SP(P_{1ij}, P_{2tk})), k = (1,2,...,N_{2t})$$

If there is no $R_{2t} = R_{1i}$, then $S_{ij} = 0$.

6) The similarity of the $i-th$ relationship in $S_1$ is

$$Sim_{ij} = \frac{\sum_{j=1}^{N_{1i}} S_{ij}}{N_{1i}}$$

7) If the reference translation is "爷爷看到了小明。" But the machine translation is "在电视上，爷爷看到了小明。" Using the above algorithm, the similarity

of each dependency relationship in $S_1$ is 1, so the algorithm needs to use $S_2$ as a reference, and repeat steps (5) (6) to get the similarity of the $i-th$ relationship in $S_2$

$$Sim_{ij} = \frac{\sum_{j=1}^{N_{2t}} S_{ij}}{N_{2i}}$$

8) The similarity between the human translation and the reference translation sentence obtained by using semantic dependency is

$$Sim_{de} = \frac{\sum_{i=1}^{n} Sim_{1i} + \sum_{i=1}^{m} Sim_{2i}}{2}$$

## 3.2 Keyword Similarity Calculation

Because the semantic similarity calculation algorithm based on semantic dependence cannot highlight keywords but calculates all dependent pairs with the same weight, some associated pairs that have little meaning for increasing semantics are also included. Keyword extraction is to structure the information contained in the text and integrate the extracted information in a unified form. Therefore, adding keyword information in this paper can supplement and optimize the semantic similarity calculation based on semantic dependence.

1) Assuming that the calculation reference translation sentence is $S_1$, and the machine translation sentence is $S_2$, call the keyword extraction interface of the semantic technology platform to obtain keyword information, and obtain the keyword and its weight set as {(Kw$_{11}$,Score$_{11}$), (Kw$_{21}$,Score$_{21}$), …, (Kw$_{n1}$,Score$_{n1}$)}, {(Kw$_{12}$,Score$_{12}$), (Kw$_{22}$,Score$_{22}$), …, (Kw$_{n2}$,Score$_{m2}$)}. Among them, $S_1$ has a total of n keywords, $S_2$ uses m keywords, $Kw_{ij}$ represents the $j-th$ $(j \leq n)$ keyword of the sentence $S_i$ $(i = 1,2)$, and Score$_{ij}$ represents the weight of the corresponding keyword. For example, "空运也面临着新的困难。" This sentence obtains the following information by calling the keyword extraction API:

Table 1 The Keywords and The Weight

| **Keywords** | 空运 | 面临 | 困难 |
|---|---|---|---|
| **Weights** | 0.751 | 0.696 | 0.602 |

As shown in Table 1, the set of keywords and their weights are {(空运,0.751), (面临,0.696), (困难,0.602)}.

2) Use the word similarity calculation method mentioned in Section 3.1 (3) to calculate the keyword similarity for the keywords obtained by $S_1$ and $S_2$, and obtain $n*m$ keyword similarity triplets $(Skw_{ij}, Score_{i1}, Score_{j2})$, where $Skw_{ij}$ represents the word similarity between the $i-th$ keyword ($Kw_{i1}$) of $S_1$ and the $j-th$ keyword ($Kw_{j2}$) of $S_2$ after keyword analysis, and the above $n*m$ triplets are summarized into one $n*m$ two-dimensional matrix.

3) The formula for calculating the keywords of two sentences is

$$Sim_{kw} = \frac{\sum_{a=1}^{\min(n,m)} K_a}{\sum_{a=1}^{\min(n,m)} weight_a}$$

Among them, $K_a$ is the $a-th$ traversal matrix to find the maximum value of $Sk_w$ in the two-dimensional matrix is $Skw_{ij}$. And $weight_a = \frac{Score_{i1}+Score_{j2}}{2}$. Each time the maximum value of similarity is found, the row and column where it is located are divided, and the next traversal is continued until the matrix is empty.

Therefore, the semantic correctness score of machine translation based on reference translation is

$$S = \frac{Sim_{kw} + Sim_{de}}{2}$$

This paper adds sentence keyword information based on semantic dependence, which can compare the semantic similarity between reference translation and machine translation more deeply and accurately from the semantic level. This results in a semantic correctness score for machine translation.

# 4 Experiment

## 4.1 Experimental Design

The corpus used in the experiment comes from English original sentences, reference translations, and five machine translation results from Google, deepl, NiuTrans, and Xinyi. Among them, 100 sets of sentences were randomly selected as the test set, and 93 sets of data, a total of 558 sentences, were obtained after cleaning the empty data, and then a machine-translated sentence with the most correct semantics in each set of sentences was manually marked. The specific method is: to perform semantic similarity calculations on each group of reference translations and five machine translations in pairs to obtain five semantic correctness scores and mark the machine translation with the highest score. Finally, the accuracy of the calculation method is judged based on the total number of sentences between the 93 sentences manually marked and the 93 sentences obtained after calculation. This article uses the accuracy rate $P$ (Precision) as the evaluation index. The accuracy rate is how many of the selected sentence samples are marked sentence samples. The specific calculation formula is:

$$P = \frac{|R_1 \cap R_2|}{R_2}$$

Among them, $R_1$ is the set of sentences marked manually, and $R_2$ is the set of sentences selected by the calculation method.

## 4.2  Results

The experimental results are shown in Table 2, where method 1 is the hownet method, method 2 uses the space vector model (VSM), method 3 is the BLEU algorithm, and method 4 is the METEOR algorithm. And "sdp" indicates the machine translation score based on semantic dependence (excluding keyword information) proposed in the article, and "Sdp+key" represents the semantic correctness evaluation algorithm of machine translation based on semantic dependence and keyword information proposed in the paper.

**Table 2 Performance Comparison of 6 Methods**

| method | *P* | method | *P* |
|---|---|---|---|
| HowNet | 0.495 | VSM | 0.559 |
| BLEU | 0.581 | METEOR | 0.602 |
| sdp | 0.655 | sdp+key | 0.795 |

## 4.3  Analysis of Results

### 4.3.1  Overall Analysis of the Algorithm

From the experimental data in the table 2, it can be seen that the calculation method in this paper has the highest accuracy rate, which is improved to a certain extent compared with similar methods. The HowNet-based calculation method only considers the similarity of word meaning, and the concepts and thesaurus are not rich enough to a certain extent; Although the BLEU algorithm has fast calculation speed and low cost, this algorithm does not consider synonyms and sentence structure, and in many cases, BLEU with the same word will give a high score; Although the METEOR algorithm takes synonyms into account, it does not consider the overall structure of the sentence; based on the space vector model, it is based on word frequency information, and the similarity between two sentences depends on the number of common words, so it cannot distinguish the semantic ambiguity of natural language. Moreover, it is assumed that words and words are independent of each other, and a keyword uniquely represents a concept or semantic unit, but the actual situation is that there are many polysemy and synonyms in the document, so this assumption is difficult to meet the actual situation.

The method in this paper calculates the similarity of the same dependency in the sentence more accurately. Although the accuracy of the algorithm for calculating the semantic similarity of sentences only through semantic dependence is improved compared with the above method, if the keyword information is added, the accuracy will decrease. Boost again. Therefore, the method in this paper considers not only the key information of sentence semantics but also the deep information of sentence semantics.

This paper mainly evaluates five machine translations according to the sentence semantic level. The evaluation of linguistic meaning mulberry is mainly represented by the calculation of the semantic similarity between keywords and texts. In the evaluation of semantics, through the calculation of the similarity of the language meaning and the matching of the language content of the reference translation text and the five machine translation texts, machine translation is considered based on the relevance, logic and consistency of the language content. How faithfully the system translates the meaning of the language.

### 4.3.2　keyword analysis

This paper uses the keyword extraction interface of the semantic technology platform to obtain keywords. These keywords are the words most related to the semantic expression of the text. These words can best reflect the theme of the text and are also the main objects of the text corpus. Taking the keyword list of the reference translation as a reference, through the comparison of the topic sub-list of the machine-translated text and the calculation of word similarity, it is found that the higher the keyword matching degree, the more complete the content conveyed by the machine-translated sentence, and the more meaningful the meaning of the text. full expression. The test set was evaluated using the keyword analysis algorithm mentioned above in this paper, and it was found that some machine-translated sentences with a keyword evaluation of less than 0.4 belonged to missing or mistranslated key information. Vocabulary omission refers to some words that should be translated but not translated in the translation.

*Case 1*：

**Original English sentence:'** *Mr. Ma has made himself scarce in public ever since.*'

**Reference Translation:**'自那以后，马云极少公开露面。'

**Newtranx:**'从那以后，马先生就一直在公共场合里变得冷淡。'

For case (1), the correctness score of this machine translation keyword is 0.290, and the keyword and the corresponding weight are obtained after the keyword is extracted from the machine translation: [{'score': '0.649', 'word': '场合'}, {'score': '0.590', 'word': '先生'}, {'score': '0.571', 'word': '冷淡'}, {'score': '0.523', 'word': '以后'}, {'score': '0.514', 'word': '公共'}, {'score': '0.503', 'word': '一直'}]. One of the key words in the sentence translated by Xinyi is "先生", which means that the literal translation of 'Mr. Ma' is '马先生', but according to the above context, it can be known that 'Mr. Ma' stands for '马云'. The other keyword "coldness" has a completely wrong meaning. It can be seen that the mistranslation of key information will lead to deviations in people's understanding of the content information.

*Case 2*：

**Original English sentence:**' Out of the top 30, 14 belong to what Chinese users call the Tencent camp. Five are in the Alibaba camp, four are owned by Baidu, and three are owned by ByteDance.'

**Reference Translation:**'在前 30 大移动应用中，有 14 个属于中国用户所说的腾讯系，五个属于阿里系，四个由百度拥有，三个由字节跳动拥有。'

**Baidu Translate:**'在前 30 名中，有 14 名属于中国用户所谓的腾讯阵营。五家在阿里巴巴阵营，四家在百度，三家在 ByteDance。'

In case (2), the correctness score of machine translation keywords is 0.320, and the keywords and corresponding weights obtained after machine translation is extracted are [{'score': '0.667', 'word': '阵营'}, {'score': '0.605', 'word': '五家'}, {'score': '0.581', 'word': '用户'}, {'score': '0.567', 'word': 'ByteDance'}, {'score': '0.554', 'word': '所谓'}, {'score': '0.552', 'word': '属于'}, {'score': '0.545', 'word': '中国'}, {'score': '0.500', 'word': '14'}, {'score': '0.500', 'word': '30'}, {'score': '0.500', 'word': '阿里巴巴'}]. Among them, "ByteDance" should be translated as "字节跳动", but Baidu Translate did not translate it, which belongs to the missing translation of vocabulary. It is found that the missing translation of vocabulary is often because the missing translation of the vocabulary is a highly professional term, and the knowledge base of machine translation This entry is not included.

*Case 3* ：

**Original English sentence:**' Mr. Bennett said that he was busy with tasks during the first part of the flight and then he heard Ms. Moses shouting, "Don't forget to look out the window."'

**Reference Translation:**' 本内特说，在飞行的第一阶段，他正忙着手头的任务，然后就听到摩西大喊，"别忘了看看窗外。"'

**Newtranx:**' 班内特先生说，在飞机的前半段，他忙于工作，然后听到摩西喊道，"别忘了往窗外看。"'

In case(3), the machine translation keyword score is 0.337, and the new translation translates "the first part of the flight" into "飞机的前半段", and the corresponding reference translation is "飞行的第一阶段". Among them, the keyword "飞行" in the reference translation is a verb, and the keyword "飞机" in the new translation is a noun. This kind of part-of-speech error seriously affects the semantic quality of machine translation. There are a large number of words with the same form but different parts of speech in both English and Chinese, but these words in English and Chinese are not completely corresponding. This difference is easier to understand and deal with in human translation, but in machine translation It is a major obstacle in translation, often generating wrong translations.

Calculate the average and variance of the keyword scores of the five machine translations to get Table 3.

Table 3 Average score and variance of five machine translation keywords

|  | Newtranx | Youdao | Baidu | Niutrans | Google |
|---|---|---|---|---|---|
| Average | 0.7446 | 0.7815 | 0.7627 | 0.7662 | 0.7866 |

| Variance | 0.0296 | 0.0218 | 0.0240 | 0.0213 | 0.0294 |

According to Table 3, it can be seen that the scores of the five machine translation keywords are in the range of 0.74-0.79, indicating that the key content of the five mainstream translation APIs on the market is relatively comprehensive. In comparison, Google Translate and Youdao The key information transmission quality of the translation is high, but the variance of the keyword score of Google Translate is large, indicating that the quality of key information between translated sentences is quite different, and the translation quality is not too stable; compared with other translation systems, the key information transmission of Xinyi Translation The quality is lower, and the quality difference between sentences is also relatively large.

### 4.3.3 Semantic Dependency Analysis

Text similarity is not only reflected in the likelihood of the combination of language fragments, but more importantly, it reflects the consistency of the meaning embodied in the language fragments. As an evaluation parameter at the semantic level, text similarity is mainly manifested in measuring the faithful effect of computers on natural language understanding and processing from the similarity between human-translated texts and machine-translated texts. This paper uses the semantic dependency analysis interface of the Semantic Technology Platform to analyze the semantic dependence of reference translations and machine translations, and calculates the semantic similarity between reference translations and machine translations based on the analyzed semantic dependencies. Semantic dependency analysis represents the semantics of a sentence based on the semantic dependency graph, so it is feasible to calculate the semantic similarity between machine translation and reference translation based on the similarity of semantic association pairs decomposed from the semantic dependency graph of two sentences.

According to the algorithm proposed above to calculate the similarity between machine translation and reference translation based on semantic dependence, we found that there are some complex and uncoordinated sentences in machine translation.

*Case 4*：

**Original English sentence:**' Others are scoffing at their bosses' return-to-office mandates and threatening to quit unless they're allowed to work wherever and whenever they want."'

**Reference Translation:**'还有人对老板重返办公室的要求嗤之以鼻，以辞职要挟老板允许他们在自己选择的时间和地点工作。'

**Baidu Translate:**'其他人则嘲笑他们的老板重返办公室，并威胁说除非他们被允许随时随地工作，否则他们将辞职。'

In example (4), the semantic similarity score obtained by Baidu Translate based on the semantic dependency analysis is 0.2119, and the semantic dependency graph obtained after the semantic analysis of the first half sentence of the reference translation is shown in Figure 1. The semantic association pairs that can reveal the main semantics of the text are ['人', '嗤之以鼻', 'Agt'] and ['要求', '嗤之以鼻', 'Datv'], where "人" is "噗之以鼻" The implementer of the sentence, and "要求" and "噗之以鼻" belong to the related relationship, to express a kind of "sniffing at the request of a certain person", and the first half of the sentence translated by Baidu is obtained according to the semantic dependency analysis The semantic dependency graph is shown in Figure 2. Among them, the semantic association pairs that can reveal the main semantics of the text are ['其他人', '嘲笑', 'Exp'] and ['老板', '嘲笑', 'Datv'], "其他人" and "嘲笑" in Baidu translation The target is not "要求" but "老板". The semantic dependency graph reveals the deep semantic relationship between the words in the sentence. It can be seen that the imprecise machine translation will lead to the semantic change of the sentence as a whole.

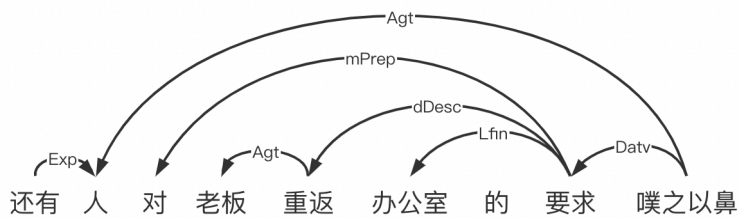

**Figure 1 The semantic dependency graph of the first half of the translation referenced in the case4**

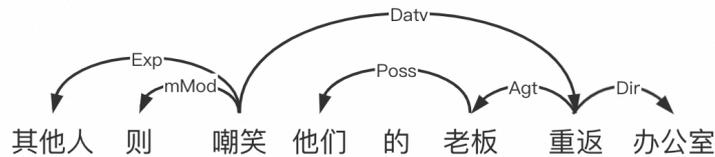

**Figure 2    The semantic dependency graph of the first half of the Baidu translate in the case4**

*Case 5*：

**Original English sentence:**' BYD, based in Shenzhen, delivered an electric S.U.V. called the Tang, to the first Norwegian customer in August.'

**Reference Translation:**'总部位于深圳的比亚迪 8 月向第一位挪威客户交付了一款名为"唐"的电动 SUV。'

**Niutrans:**'总部位于深圳的比亚迪交付了一款电动 S.U.V 给唐先生打电话，给八月份的第一个挪威客户。'

In example (5), "called the Tang" in the original sentence should be translated as "名为'唐'" according to the whole sentence, but the direct translation in Niutrans is "给唐先生打电话", this kind of error belongs to translation Wrong choice of word, "called" can be translated into "被称为" or "打电话". According to the context and meaning of the sentence, the semantic meaning of "被称为" should be selected. Niutrans 's wrong translation choice caused a serious sentence semantic error. Not only is the semantic error wrong, but it is also very complicated to read. We take the semantic dependency graph of Niutrans' translated sentence as "," as shown in Figure 3 for the first half of the sentence. Observe the semantic dependency graph ['SUV', '打电话', 'Tool '] means that the tool for "打电话" is "SUV", which can also be judged to be a wrong translation according to readers' common sense, ['交付', '打电话', 'ePurp'] where "ePurp" indicates the purpose relationship, so this semantic The associated pair indicates that the purpose of "交付" is to "打电话", while the original English sentence has no expression of this meaning at all. This mistranslation can cause confusion for the reader and affect the understanding of the context.

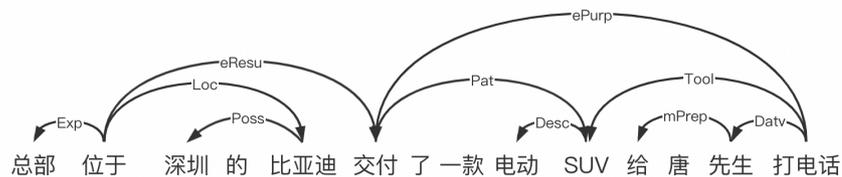

Figure 3 The semantic dependency graph of the first half of the Newtranx in the case4

*Case 6*：

**Original English sentence:**' Her credibility has been enhanced because she also operates as a creator.'

**Reference Translation:**' 由于她也是创作者，这增加了她的可信度。'
**Niutrans:**' 她的可信度提高了，因为她也作为创作者运作。'
**Google:**' 她的可信度得到了提高，因为她还以创作者的身份运作。'

In example (6), we will analyze the semantic dependency between the reference translation and Niutrans, and the obtained semantic dependency graphs are shown in Figure 4 and Figure 5. According to the semantic dependency graph, the backbone of the reference translation "Because she is also the creator" For "she is the creator", ['她', '运作', 'Agt'] in Niutrans means that "她" is the implementer of "运作", ['作为', '创作者', 'mPrep'] where "mPrep" is a prepositional marker so "as a creator" is a prepositional phrase, so the backbone of "因为她也作为创作者运作" in Niutrans is "她运作". The backbone of the reference translation and the Niutrans are completely different, and readers feel a bit complicated when reading the sentences translated by Niutrans in Example (6). The reason is because 'operates as a creator' in the original sentence should be translated as "作为一个创造者" or "是创作者", in which 'operates as' is a verb phrase composed of "verb + preposition" which means "作为" or " 是" and Niutrans directly disassembled this phrase and directly translated 'operates' into "运作", and the same is true for Google Translate in example (6). Although the overall semantics of the sentence are not affected too much, readers read it It will feel complicated and incoherent.

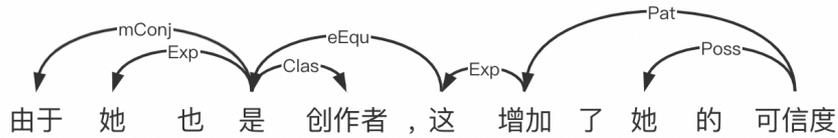

**Figure 4 Semantic dependency graph for reference translation in case6**

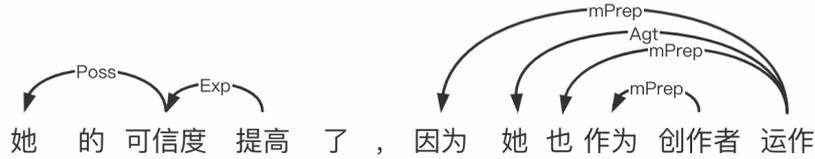

**Figure 5 Semantic dependency graph for Niutrans in case6**

The mean value and variance of the semantic scores calculated by the five machine translations based on the semantic dependency analysis (excluding keyword information) are shown in Table 4.

**Table 4 The mean and variance of five machine translation semantic dependency analysis scores**

|  | Newtranx | Youdao | Baidu | Niutrans | Google |
|---|---|---|---|---|---|
| **Average** | 0.5240 | 0.5780 | 0.5740 | 0.5684 | 0.5938 |
| **Variance** | 0.0219 | 0.0310 | 0.0284 | 0.0278 | 0.0415 |

According to Table 4, it can be seen that the average scores of the five machine translations according to the semantic dependency analysis are all between 0.5-0.6, and Google Translate has the highest score, but the variance is also the largest, indicating that Google Translate is generally of high quality, but there are still many inappropriate erroneous translations; the translation quality of the Newtranx is slightly lower than the semantic quality of the other four machine translation translations.

# 5  Conclusion

The method proposed in this paper to evaluate the semantic correctness of machine translation based on semantic dependency and keyword information starts from the semantic dependency graph of the sentence to extract the dependency relationship, dependent words, and dependent words, and then takes into account the deep semantic dependency of the sentence; And through the keyword information, considering the key information that affects the semantics of the sentence, the method of combining the key information and the semantic dependency information can comprehensively evaluate the semantic correctness of the machine translation sentence. Experiments show that the accuracy of the algorithm has been improved compared with similar methods, which proves its effectiveness, but does not consider the authenticity of grammatical information and machine-translated sentences. The next step is to obtain the machine translation evaluation score based on the dependent grammar information, and add it to the machine translation evaluation calculation.